# Clustered Bandits

Loc Bui, Ramesh Johari, and Shie Mannor

**Abstract**

We consider a multi-armed bandit setting that is inspired by real-world applications in e-commerce. In our setting, there are a few types of users, each with a specific response to the different arms. When a user enters the system, his type is unknown to the decision maker. The decision maker can either treat each user separately ignoring the previously observed users, or can attempt to take advantage of knowing that only few types exist and cluster the users according to their response to the arms. We devise algorithms that combine the usual exploration-exploitation tradeoff with clustering of users and demonstrate the value of clustering. In the process of developing algorithms for the clustered setting, we propose and analyze simple algorithms for the setup where a decision maker knows that a user belongs to one of few types, but does not know which one.

## I. INTRODUCTION

*Multi-armed bandit* (MAB) models are a benchmark model for learning to make decisions under uncertainty. In the classical stochastic model [5], [4], a decision maker chooses between several alternatives ("arms") that offer uncertain and unknown payoff; through successive experimentation ("exploration") on the arms the decision maker learns those alternatives that are most valuable, and proceeds to use those in the future ("exploitation"). MAB models are used in a very wide range of application areas featuring stochastic decision making, including pricing, marketing, advertising, product selection, and recommendation systems.

In many application areas of interest for MAB models, however, the decision maker faces two simultaneous challenges. First, the decisions made may be *context-specific*. For example, in online recommendation systems (such as the one used by Amazon), the product recommended to a given user (the "arm") depends on the characteristics of the user herself: her demographics, past purchase history, etc.

L. Bui is with the School of Engineering, Tan Tao University, Vietnam (e-mail: locbui@ieee.org). R. Johari is with the Department of Management Science and Engineering, Stanford University, USA (e-mail: ramesh.johari@stanford.edu). S. Mannor is with the Department of Electrical Engineering, Technion, Israel (e-mail: shie@ee.technion.ac.il).

In response to this challenge, recent literature has considered a *contextual* version of the classical MAB model [6], [3], [8], [7].

The second challenge is that in general, the number of contexts may be quite large, and the number of observations per context may be quite small; this is certainly the case in most recommendation settings, where each user may only be interacting with the system for a relatively small number of purchases, and yet the number of users can be quite high. In these settings the only way for the decisions maker to effectively learn is to exploit latent *low-dimensional* structure in the high dimensional problem; that is, to identify a few features that capture most of the heterogeneity of contexts, or to cluster the contexts into a few groups with similar characteristics.

In practice, these two challenges are addressed in separate phases. Typically, the decision maker estimates low-dimensional structure from high-dimensional data *offline*; i.e., based on previously collected data about contexts. After doing this estimation, the inferred low-dimensional representation is used in solving the online contextual bandit problem [10], [9]. In other words, exploration and exploitation in real time is restricted to learning only how a given context fits into the previously inferred low dimensional structure; the low-dimensional representation is only refined on much longer timescales, and is effectively decoupled from learning.

In this paper we propose a model that combines both low-dimensional estimation and online learning, that we refer to as *clustered bandits*. The main motivation is that by combining the two, exploration can be made more intelligent. In particular, we account for the benefit not only of learning the correct decision for a particular context, but also how more information about a context informs the low-dimensional representation of the overall context space.

In our model, we assume that users ("contexts") arrive over time, and the decision maker must choose the best decision for each user, from a fixed finite set of alternatives. Though there may be a large number of users, we assume that users come from only a fixed set of finitely many *types*; users of the same type give the same average reward on each arm. The decision maker, of course, does not observe the true type of each user, and does not know the average reward vector of each type. Thus the decision maker must simultaneously cluster the users into groups by type, and also make the best decision for each type.

Our main contributions are as follows. *First*, we propose a novel model of multiarmed bandits, incorporating the combination of low-dimensional estimation and online learning described above. This model lays the foundation for analysis of combined estimation and bandit problems that is of critical importance for practical applications. *Second*, we propose two distinct approaches to developing algorithms for this setting: one where we first explore, then estimate clusters, then exploit; and another where we continuously cluster. We provide novel algorithms in each setting, as well as analysis of regret performance. *Third*, we use numerical experiments to demonstrate the value of clustering over time.

## II. PROBLEM FORMULATION

We consider the problem of stochastic multi-user multi-armed bandits, called *clustered bandits*. The setting is as follows. There are a finite set of users $\mathcal{U} = \{1, 2, \ldots, M\}$ and a finite set of user types $\mathcal{X} = \{1, 2, \ldots, N\}$. We assume that every user belongs to some type, and we are interested in the case where $M$ is generally much larger than $N$. At each time $t = 1, 2, \ldots, T$, a user $u \in \mathcal{U}$ arrives into an online system, then the system has to choose an action (arm) $a \in \mathcal{A}$, where $\mathcal{A} = \{1, 2, \ldots, K\}$ is the set of arms; upon choosing the arm, the system obtains a reward. We assume i.i.d. Bernoulli rewards, whose expectations depend on both the action taken and the type of that user. In particular, let $x$ be the current user's type; then the expected reward of action $a$ is denoted by $\theta_x(a)$. For convenience, let $\vec{\theta}_x$ denote the expected reward vector under type $x$, i.e.,

$$\vec{\theta}_x := [\theta_x(1), \theta_x(2), \ldots, \theta_x(K)] \in [0, 1]^K.$$

Also, define:

$$\theta_x^* := \max_{1 \le a \le K} \theta_x(a), \quad a_x^* := \arg\max_{1 \le a \le K} \theta_x(a), \quad \Delta_x(a) := \theta_x^* - \theta_x(a).$$

For simplicity in our paper, we assume that $a_x^*$ is unique for each $x$, though our results extend naturally even without this assumption. We emphasize that an important aspect of this problem is that for all users belonging to the same type, say $x$, the expected reward vector across the arms is *exactly the same*, given by $\vec{\theta}_x$. We also call $\vec{\theta}_x$ a *parameter vector*, and $\{\vec{\theta}_x\}_{x \in \mathcal{X}}$ the *parameter set*.

Our goal is to propose a policy (algorithm) to take an action (pull an arm) whenever a user arrives. We assume that the system *only* has access to the following information: number of actions ($K$), number of types ($N$), and the user's ID when he or she arrives ($u$). Note that we assume the number of types is known; this assumption is made for simplicity, and corresponds to an *ex ante* determination of the number of "clusters" of interest to the decision maker. More generally, an interesting open direction concerns online identification of the right number of clusters.

The performance of a policy is measured as regret with respect to the *oracle policy* that knows both the type of each user and the reward vector under each type. In particular, let $u^t$ be the user arriving at time $t$, $x(u^t)$ be the type of user $u^t$, and $a^t$ be the action taken by the policy at time $t$. Then the expected regret is defined as:

$$\mathbb{E}[\mathsf{Reg}] = \sum_{t=1}^{T} \theta_{x(u^t)}^* - \sum_{t=1}^{T} \theta_{x(u^t)}(a^t).$$

*A. Discussion of existing solutions*

Since each user corresponds to a stochastic MAB problem, one naive approach is to treat users separately, and run a separate stochastic MAB algorithm, e.g., UCB (Upper Confidence Bound) [2] or SE (Successive Elimination) [7], for each of them. The expected regret in this case is $O(M \ln(T/M))$. However, this approach does not take into account the important feature that many users may have the *same* expected reward vector. In particular, if the number of users $M$ is much larger than the number of types $N$, treating each user separately is very inefficient because it fails to exploit latent low-diemensional structure.

As mentioned in the Introduction section, this problem is also very similar to the *contextual bandit* problem [6], [3], [8], [7], in which the user's ID is treated as the "context." However, the existing solutions for contextual bandits do not seem to be applicable to this setting. In particular, solutions in [6], [3] involve finding the best policy among a limited set of policies mapping from the context (i.e., the user's ID in this case) to the action. On the other hand, [8], [7] consider the scenario in which each context is assumed to be associated with an arm configuration, and hence, their solutions are the same as the per-user stochastic MAB approach above.

*B. Our approaches*

Our main idea is to exploit the *latent low-dimensional structure*: there are only a few types of users, and the expected rewards from users belonging to the same type are exactly the same. This suggests if we can "group" the users that belong to the same type to the same group, then we can leverage the latent low-dimensional type information despite the potentially high number of users. To this end, we consider two different approaches in this paper, described below.

**Approach 1: Exploration, clustering, exploitation.** First, we assume that we know the parameter set $\{\vec{\theta}_x\}_{x \in \mathcal{X}}$, but not the exact type of each arriving user. We discuss an algorithm that can efficiently solve this problem for one user, and inspired by this algorithm, demonstrate that the following approach can be used: first, we explore over an initial set of users; then we cluster from this data; and finally we run that algorithm for new users. This corresponds closely to actual practice in collaborative filtering systems.

This setting uses as a subroutine an efficient algorithm in the following setting: The system knows the set of parameter vectors, but does not know which one is the *true* one. Since the system has "some" information about the rewards, it is expected that the regret will be better than the traditional multi-armed bandit setting. This setting is addressed with more details in the next section.

**Approach 2: Online clustering.** The previous approach does not include clustering as an online component of the learning process: it is executed once and set for the remainder of the decision horizon. An alternative is to continuously cluster as we learn. In particular, suppose first that we know the type of each arriving user (but do not know the parameter vector for each type); then the problem is reduced to $N$ separate bandit problems, one for each type. Thus a reasonable approach including online clustering proceeds as follows: In each time step, given past observations, we use a clustering technique to divide users into $N$ groups. Then we treat the problem as if the types of users are known.

## III. Detour: (single user) multi-armed bandits with known parameter set

We make a detour in this section by considering a variation of the standard MAB problem. The variation itself is of independent interest, but more importantly, the results (and algorithms) for this setting will be used to elaborate the exploration-clustering-exploitation approach described above to tackle the clustered bandit problem.

The detailed setting is as follows. There is a finite set of parameters $\mathcal{X} = \{1, 2, \ldots, N\}$, a finite set of arms $\mathcal{A} = \{1, 2, \ldots, K\}$, and a finite set of reward vectors $\{\vec{\theta}_z\}_{z \in \mathcal{X}}$; all are *known* to the system. However, the *true* rewards are driven by an *unknown* parameter $x \in \mathcal{X}$ (i.e., the true reward vector is $\vec{\theta}_x$, although the system knows that $x \in \mathcal{X}$). Note that we are not considering multiple users here.

The performance of a policy for this problem is measured as regret with respect to the policy that knows the true parameter. In particular, if $x$ is the true parameter, and $a^t$ is the action taken at time $t$, then the expected regret is defined as

$$\mathbb{E}[\mathsf{Reg}] = T\theta_x^* - \sum_{t=1}^{T} \theta_x(a^t).$$

This problem is different from the traditional MAB problem in the sense that the system has "some" information about the expected rewards: it is known that the reward vector belongs to a known set. Of course, one can ignore that fact and apply a UCB-like algorithm naively (to get an expected regret of $O(K \ln T)$). A more clever approach can yield the regret $O(\min\{N, K\} \ln T)$ by considering only the set of optimal arms (each corresponding to a parameter) when $N < K$. But can we do better than that? In the next subsection, we will present an algorithm that, depending on the structure of parameter set, can achieve $O(1)$ regret in some cases.

### A. The UCB-KT (Upper Confidence Bound with Known Types) algorithm

Let $\bar{\theta}_t$ be the empirical reward vector up to time $t$. Given a parameter $x$, define $B(x)$ as the set of parameters for which the optimal arms are better than the optimal arm for $x$, and the reward distribution under the optimal arm for $x$ are the same, i.e.,

$$B(x) := \{z \in \mathcal{X} : \theta_z(a_x^*) = \theta_x(a_x^*), a_z^* \neq a_x^*\}.$$

Also, let $\mathcal{E}$ denote the set of "elite" arms, i.e., $\mathcal{E} := \{i \in \mathcal{A} : i = a_x^* \text{ for some } x \in \mathcal{X}\}$. That is, $\mathcal{E}$ includes only arms that are optimal for some parameter $x \in \mathcal{X}$. Note that $|\mathcal{E}| \leq |\mathcal{X}| = N$.

For a value $\theta \in [0, 1]$, let us define the $\epsilon$-neighborhood of $\theta$ as follows.

$$\epsilon\text{-nbd}(\theta) := \{\mu \in [0, 1] : |\mu - \theta| < \epsilon\}.$$

Since $\mathcal{X}$ is finite, it is possible to find an $\epsilon^* > 0$ such that all $\epsilon^*$-neighborhoods of $\theta_x(a)$ are disjoint for *distinct* values of $\theta_x(a)$, $x \in \mathcal{X}$, $a \in \mathcal{A}$.

Now, let us define the following conditions:

- $C1(x)$: $\bar{\theta}_t(a) \in \epsilon^*\text{-nbd}(\theta_x(a))$, $\forall a \in \mathcal{A}$, and $B(x)$ is *empty*;
- $C2(x)$: $\bar{\theta}_t(a) \in \epsilon^*\text{-nbd}(\theta_x(a))$, $\forall a \in \mathcal{A}$, and $B(x)$ is *non-empty*;
- $C3$: there does not exist $x \in \mathcal{X}$ such that $\bar{\theta}_t(a) \in \epsilon^*\text{-nbd}(\theta_x(a))$, $\forall a \in \mathcal{A}$.

The UCB-KT algorithm is as follows (its detailed pseudo-code is presented in the Appendix A). First, identify the value of $\epsilon^*$ based on the parameter set $\{\vec{\theta}_x\}_{x \in \mathcal{X}}$. Then, for $t = 1, \ldots, K$, pull each arm once. For each $t \geq K + 1$:

- If $C1(x)$ is satisfied for some $x \in \mathcal{X}$, pull $a_x^*$ (the optimal arm under $x$);
- If $C2(x)$ is satisfied for some $x \in \mathcal{X}$, perform one step of the UCB algorithm [2] on the set of "elite" arms $\mathcal{E}$, i.e., pull the arm that achieves:

$$\arg\max_{a \in \mathcal{E}} \bar{\theta}_t(a) + \sqrt{\frac{2 \ln(t)}{T_t(a)}},$$

where $T_t(a)$ is the number of times that arm $a$ has been pulled up to time $t$;
- If $C3$ is satisfied, round-robin among all arms.

The main idea is that even if round-robin ($C3$) moves us to the wrong $x$, if $B(x)$ is empty, then pulling the optimal arm under $x$ ($C1$) will eventually move our parameter estimate away from $x$. However, if the round-robin in ($C3$) moves us to the wrong $x$, and $B(x)$ is *non-empty*, then pulling only $a_x^*$ does not help us at all: it cannot move us away from $x$ when the true parameter is in $B(x)$, because under $a_x^*$ both the wrong and true parameters yield the same reward. Thus, we need to explore more on $\mathcal{E}$ and in the proposed algorithm we use UCB to do that.

*Theorem 1 (Upper bound for UCB-KT):* Let $x$ be the true parameter. Then there exists a constant $\gamma_A > 0$, which depends on $\epsilon^*$, such that for the UCB-KT algorithm:

- if $B(x)$ is empty, then

$$\mathbb{E}[\mathsf{Reg}]_{\text{UCB-KT}}(x, T) \leq \gamma_A;$$

- if $B(x)$ is non-empty, then

$$\mathbb{E}[\mathsf{Reg}]_{\text{UCB-KT}}(x, T) \leq \sum_{a \in \mathcal{E}} \frac{8}{\Delta_x(a)} \ln T + \gamma_A.$$

The proof is presented in the Appendix B. The theorem says that the UCB-KT's performance depends on the structure of the parameter set and the true parameter. In particular, if $B(x)$ is non-empty for the true parameter $x$, then the regret of UCB-KT is $O(\min\{N, K\} \ln T)$, which is the same as the UCB-based approach mentioned above. However, if the set $B(x)$ is empty for the true parameter $x$, then UCB-KT can achieve $O(1)$ regret.

One may ask if we can do better than $O(\ln T)$ when $B(x)$ is non-empty for the true parameter $x$. The answer turns out to be negative. In particular, it has been proved by Agrawal et al. [1] that, if $x$ is the true parameter and the set $B(x)$ is non-empty, under some mild conditions, the regret of *any* algorithm $\phi$ satisfies the following:

$$\liminf_{T \to \infty} \frac{\mathbb{E}[\mathsf{Reg}]_\phi(x, T)}{\ln T} \geq \min_{\alpha \in P_{-x}} \max_{z \in B(x)} \frac{\sum_{\mathcal{A}_{-x}} \alpha(a) \Delta_x(a)}{\sum_{\mathcal{A}_{-x}} \alpha(a) I^a(x \| z)}, \tag{1}$$

where $\mathcal{A}_{-x} := \mathcal{A} \setminus \{a_x^*\}$ is the set of all arms except the optimal arm under $x$, $P_{-x}$ is the simplex over $\mathcal{A}_{-x}$, and $I^a(x \| z)$ is the Kullback-Leibler divergence between reward distributions of arm $a$ under $x$ and $z$.

We conclude this subsection by noting that Agrawal et al. [1] also proposed an algorithm for this setting, and proved that their algorithm can match the lower bound in (1). However, their algorithm is rather complicated, and more importantly, requires precomputation of the optimal distribution $\alpha^*$ that achieves the minimization in (1). This requirement turns out to be a crucial point that makes their algorithm inapplicable for the clustered bandit problem. On the other hand, the proposed UCB-KT algorithm is simpler, offers a competitive performance, and as we will show later, can be modified to apply to the clustered bandit setting.

### B. Numerical results

In this section, we perform a numerical experiment to verify the performance of UCB-KT. The experiment includes a parameter set of $N = 21$ parameter vectors, indexed from 0 to 20. The number of arms is $K = 21$ (also indexed from 0 to 20). For parameter $x = 0$, we have that $\theta_0(0) = 0.55$ and $\theta_0(a) = 0.5$ for $a = 1, \dots, 20$. For parameters $x = 1, \dots, 20$, we have that $\theta_x(0) = 0.55$, $\theta_x(x) = 0.6$, and $\theta_x(a) = 0.5$ for $a \neq x$. Therefore, $B(x = 0)$ is non-empty, while $B(x)$ is empty for $x \neq 0$.

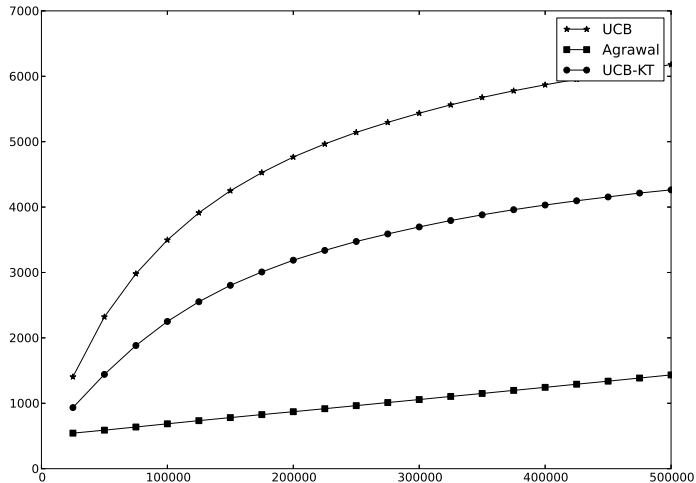

Fig. 1. UCB vs. UCB-KT vs. Agrawal et al.

We run the experiment for 100 runs. For each run, we pick $x = 0$ as the true parameter with probability $1/2$, while picking each of the rest as the true parameter with probability $1/(2 * 20)$. (In this way, $B(x)$ is empty for roughly half of the runs.) Figure 1 shows the average regrets of three algorithms: UCB (on the set of optimal arms), UCB-KT, and Agrawal et al. (the optimal distribution $\alpha^*$ for this parameter set can be pre-computed easily, and we omitted the error bars because the variation was small). One can see that UCB-KT performs better than UCB but not as good as the Agrawal et al. algorithm, which is expected from the theory.

## C. The UCB-KT($\delta$) algorithm

In this last part of the detour, let us present a variation of the UCB-KT algorithm, called UCB-KT($\delta$), in which we introduce some disturbance to the set $B(x)$. This variation will be useful for the development of clustered bandit algorithms in the next section.

**UCB-KT($\delta$):** same as the UCB-KT algorithm, but using the following $B(x, \delta)$ instead of $B(x)$:

$$B(x, \delta) = \{z : \text{for some } a' \text{ s.t. } \theta_x(a') \geq \sup_a \theta_x(a) - 2\delta, \text{ there holds } |\theta_z(a') - \theta_x(a')| \leq 2\delta,$$

$$\text{while there is at least one } a \neq a' \text{ such that } \theta_z(a) > \theta_z(a') - 2\delta\}.$$

*Theorem 2 (Upper bound for UCB-KT($\delta$)):* Let $x$ be the true parameter. Then there exists a constant $\gamma_A > 0$, which depends on $\epsilon^*$, such that for the UCB-KT($\delta$) algorithm:

- if $B(x, \delta)$ is empty, then

$$\mathbb{E}[\mathsf{Reg}]_{\text{UCB-KT}(\delta)}(x, T) \leq \gamma_A;$$

- if $B(x, \delta)$ is non-empty, then

$$\mathbb{E}[\mathsf{Reg}]_{\text{UCB-KT}(\delta)}(x, T) \leq \sum_{a \in \mathcal{E}} \frac{8}{\Delta_x(a)} \ln T + \gamma_A.$$

Note that $B(x) \subseteq B(x, \delta)$ for any $\delta > 0$. Thus, if $B(x, \delta)$ is empty, $B(x)$ is also empty. The proof then follows the same steps as the proof of Theorem 1. We omit the details.

## IV. Algorithms for clustered bandits

Based on the ideas and results developed before, in this section, we propose three algorithms for the clustered bandit problem. The first two algorithms are based on the exploration-clustering-exploitation approach and the results obtained in Section III, while the last algorithm is based on the online clustering approach.

### A. Unif - Clustering - UCB-ET($\delta$)

This algorithm has an input parameter $M_0$ and works as follows: For the first $M_0$ users (called *pilot users*), sample the arms uniformly at random. After a large enough number of pilot users, perform a clustering algorithm (with $N$ being the number of clusters) to obtain $N$ *estimated* parameter vectors $\{\bar{\theta}_x\}$. Then, run the UCB-ET($\delta$) algorithm for the new (non-pilot) users.

**UCB-ET($\delta$) (UCB with Estimated Types):** same as the UCB-KT($\delta$) algorithm, but using the following $\hat{B}(x, \delta)$ instead of $B(x, \delta)$. (The only difference between them is that $\hat{B}(x, \delta)$ is defined on the estimated parameter vectors $\{\bar{\theta}_x\}$ instead of the true parameter vectors $\{\theta_x\}$.)

$$\hat{B}(x, \delta) = \{z : \text{for some } a' \text{ s.t. } \bar{\theta}_x(a') \geq \sup_a \bar{\theta}_x(a) - 2\delta, \text{ there holds } |\bar{\theta}_z(a') - \bar{\theta}_x(a')| \leq 2\delta,$$

$$\text{while there is at least one } a \neq a' \text{ such that } \bar{\theta}_z(a) > \bar{\theta}_z(a') - 2\delta\}.$$

The detailed pseudo-code of the algorithm is presented in the Appendix C.

*B. UCB - Clustering - UCB-ET(δ)*

This algorithm is the same as the previous algorithm except for one point: for the first $M_0$ users, we sample arms according to the UCB policy rather than uniformly. The detailed pseudo-code of the algorithm is presented in the Appendix D.

This algorithm and the previous algorithm illustrate why the algorithm by Agrawal et al. is not applicable for this approach to the clustered bandit problem: running the algorithm by Agrawal et al. requires the computation of the optimal distribution $\alpha^*$ (that achieves the minimization in (1)) on the *estimated* parameter vectors, which cannot be precomputed (since the estimated parameter vectors are *ex ante* random).

*C. Continuous clustering & UCB*

This algorithm is based on an online clustering approach (Section II-B): at *every* time slot, perform a clustering algorithm (with $N$ being the number of clusters) on *all* users' empirical reward vectors to divide them into $N$ groups. We then run $N$ separate UCB policies, one per group; i.e., at the current time step, we then take an action according to the UCB policy specialized to the group containing the current user. The detailed pseudo-code of the algorithm is presented in the Appendix E.

## V. NUMERICAL RESULTS

In this section, we perform a numerical experiment to evaluate the performance of the proposed algorithms for clustered bandits. The clustering method being used is $k$-means. The experiment includes a parameter set of $N = 2$ parameter vectors, each with $K = 4$ arms: $\theta_0 = [0.6, 0.5, 0.5, 0.5]$ and $\theta_1 = [0.5, 0.6, 0.5, 0.5]$. (That is, both $B(0)$ and $B(1)$ are empty.) In total, there are 2000 users arriving over time, and each of them stays for exactly $\tau = 100$ time slots. Each arriving user is of type $0$ with probability $1/2$ and of type $1$ with probability $1/2$.

Figure 2 shows the average regrets of various algorithms: **UCB-per-user**, $k$**-means & UCB continuously**, **Unif -** $k$**-means - UCB-ET(δ)** with $M_0 = 40$ and $\delta = 0.01$, **UCB -** $k$**-means - UCB-ET(δ)** with $M_0 = 40$ and $\delta = 0.01$, **Agrawal et al.**, and **UCB-on-types** (again, we omitted the error bars because the variation was small). Note that the **Agrawal et al.** algorithm requires the parameter set to be known in advance, and the **UCB-on-types** algorithm requires the type of each user to be known in advance. Thus, the **Agrawal et al.** serves as the lower bound for the **Unif -** $k$**-means - UCB-ET(δ)** and the **UCB -** $k$**-means - UCB-ET(δ)**, while the **UCB-on-types** serves as the lower bound of $k$**-means &**

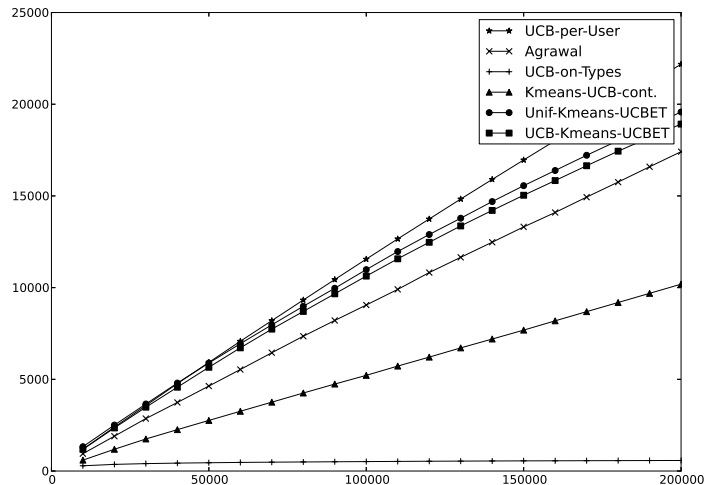

Fig. 2.   Regrets of various algorithms for clustered bandits

**UCB continuously**. We can see that all the proposed algorithms perform better than the **UCB-per-user** algorithm, and particularly, the $k$-**means & UCB continuously** works extremely well.

The preceding insight is an important result of our paper. In particular, this demonstrates the value of online clustering in such settings: the decision maker can do significantly better through more frequent re-estimation of the latent low-dimensional structure in the system.

## VI. THEORETICAL ANALYSIS

In this section, we outline an approach to regret bounds for algorithms that jointly cluster and learn. Our approach consists of two steps. First, we assume that the clustering algorithm we use can be characterized by an error probability that depends on the number of samples observed and the desired confidence. Next, we demonstrate how this performance can be coupled to our earlier theoretical analysis of UCB-KT($\delta$) to obtain regret bounds.

For this section, for concreteness, we consider the Unif-Clustering-UCB-ET($\delta$) scheme discussed in Section IV-A. For purposes of a theoretical model, we assume that users arrive over time, stay for exactly $\tau$ time slots, and then leave. Further, we assume that each user is sampled uniformly at random from the parameter set $\mathcal{X}$.

Recall that Unif-Clustering-UCB-ET($\delta$) works by using a clustering scheme after the first $M_0$ users are sampled. We start with the following assumption on this clustering scheme.

*Assumption 1 (Hypothetical clustering scheme):* Fix $\delta > 0$ and $M_0$. Suppose that for the first $M_0$ users, we sample arms uniformly at random at every time step from the arm set $\mathcal{A}$; we then cluster the users into $N$ clusters based on their empirical average reward vectors. Let $\bar{\theta}_x(a)$ be the empirical average reward of users assigned to cluster $x$. We assume the clustering algorithm is such that:

$$\mathbb{P}(\min_\sigma \max_{x,a} |\bar{\theta}_{\sigma(x)}(a) - \theta_x(a)| \geq \delta) \leq g(\delta, M_0),$$

for some function $g$ that is increasing in $\delta$ and decreasing in $M_0$. (Here $\sigma$ ranges over all permutations of $\mathcal{X}$.)

The previous assumption says that (up to a relabeling of the empirical reward vectors for each cluster), our clustering scheme is able to identify the correct clusters with high probability. Now observe that on the complement of the event identified in the preceding assumption, we will have learned the correct cluster centers to within confidence $\delta$. Thus on this event we expect that UCB-ET($\delta$) will perform well. On the event in the preceding assumption, however, our regret can be as bad as linear. Further, as we lower $\delta$, we make $\hat{B}(x, \delta)$ smaller, but at the expense of higher potential error in clustering. Thus the tradeoff is between the high regret when we fail to cluster effectively, and the high regret when we cluster to within a $\delta$ bound but $\hat{B}(x, \delta)$ is too large to give low regret.

With this in mind, consider the Unif-Clustering-UCB-ET($\delta$) scheme in which we do uniform sampling for the first $M_0$, then cluster by the hypothetical clustering algorithm in Assumption 1, and then run UCB-ET($\delta$) for the remaining $M - M_0$ users. Then the regret of this scheme is upper bounded as follows.

*Theorem 3:* Suppose that users arrive sequentially over time, sampled uniformly at random from the parameter set $\mathcal{X}$, and stay for $\tau$ periods each before leaving. The expected regret of the above Unif-Clustering-UCB-ET($\delta$) scheme is upper bounded in the following result.

$$\mathbb{E}[\mathsf{Reg}] \leq M_0 \sum_{x \in \mathcal{X}} \frac{1}{N} \sum_{a \in \mathcal{A}} \frac{\Delta_a(x)\tau}{K} + g(\delta, M_0)(T - M_0\tau)(\max_{x,a} \Delta_a(x))$$

$$+ (1 - g(\delta, M_0)) \left(\frac{T}{\tau} - M_0\right) \sum_{x \in \mathcal{X}} \frac{1}{N} \mathbb{E}[\mathsf{Reg}]_{\text{UCB-KT}(2\delta)}(x, \tau),$$

where UCB-KT($\delta$) is as defined in Section III-C.

The proof of the theorem involves showing that the regret of UCB-ET($\delta$) is bounded above by the regret of UCB-KT($2\delta$). The result follows if on the complement of the event in Assumption 1, we have $\hat{B}(x, \delta) \subseteq B(x, 2\delta)$; and this follows by a straightforward calculation from the definitions. We omit the details.

The preceding theorem formalizes the tradeoff discussed above. In particular, in optimizing the regret bound, two parameters are considered. First, by increasing $M_0$, we obtain better confidence in our clustering (and thus a smaller $g(\delta, M_0)$, at the expense of high regret in the initial phase. Second, by increasing $\delta$, we also obtain a smaller $g(\delta, M_0)$, but at the expense of potentially larger sets $B(x, 2\delta)$, and thus higher regret in the final phase. For a given clustering scheme (and thus a given $g$, minimizing over $M_0$ and $\delta$ would yield the best regret bound of this type.

APPENDIX

## A. The UCB-KT algorithm

---

**Algorithm 1** UCB-KT

---

**Input:** $t$ (current time), $K$ (number of arms), $\{\vec{\theta}_z\}_{z \in \mathcal{X}}$ (set of reward vectors), $\epsilon^*$ (parameter)

**Output:** $I_t$ (index of the arm to be pulled)

**if** $t \leq K$ **then**

$\quad I_t \leftarrow t$ (pull each arm once)

**else**

$\quad$ **if** $C1(x)$ is satisfied for some $x \in \mathcal{X}$ **then**

$\qquad I_t \leftarrow a_x^*$ (the optimal arm under $x$)

$\quad$ **else if** $C2(x)$ is satisfied for some $x \in \mathcal{X}$ **then**

$\qquad I_t \leftarrow \arg\max_{a \in \mathcal{E}} \bar{\theta}_t(a) + \sqrt{\dfrac{2\ln(t)}{T_t(a)}}$ (UCB on the set of "elite" arms $\mathcal{E}$)

$\quad$ **else**

$\qquad I_t \leftarrow ((t-1) \mod K) + 1$ (round-robin among all arms)

$\quad$ **end if**

**end if**

---

## B. Theorem 1 and its proof

First, we prove the following lemma.

*Lemma 1:* Let $X_1, X_2, \ldots, X_n$ be i.i.d. Bernoulli random variables with mean $\mu$. Let

$$\overline{\mu}_n := \frac{1}{n} \sum_{i=1}^{n} X_i, \quad \text{and} \quad L_\epsilon = \sup\{n \geq 1 : |\overline{\mu}_n - \mu| \geq \epsilon\},$$

for some $\epsilon > 0$. Then,

$$\mathbb{E}[L_\epsilon] \leq \gamma(\epsilon), \quad \text{where} \quad \gamma(\epsilon) := \frac{2e^{-2\epsilon^2}}{(1 - e^{-2\epsilon^2})^2}.$$

*Proof:* By the Chernoff-Hoeffding bound, we have that:

$$\mathbb{P}(|\overline{\mu}_n - \mu| \geq \epsilon) \leq 2e^{-2\epsilon^2 n}.$$

Note that

$$
\begin{aligned}
\mathbb{E}[L_\epsilon] &= \mathbb{E}\left[\sum_{n=1}^{\infty} \mathbf{1}\left(\exists i \geq n : |\overline{\mu}_i - \mu| \geq \epsilon\right)\right] = \mathbb{E}\left[\sum_{n=1}^{\infty} \mathbf{1}\left(\bigcup_{i \geq n}\left(|\overline{\mu}_i - \mu| \geq \epsilon\right)\right)\right] \\
&\leq \sum_{n=1}^{\infty}\sum_{i=n}^{\infty} \mathbb{P}\left(|\overline{\mu}_i - \mu| \geq \epsilon\right) \leq 2\sum_{n=1}^{\infty}\sum_{i=n}^{\infty} e^{-2\epsilon^2 i} \\
&= 2\sum_{n=1}^{\infty} \frac{e^{-2\epsilon^2 n}}{1 - e^{-2\epsilon^2}} = \frac{2e^{-2\epsilon^2}}{(1 - e^{-2\epsilon^2})^2}.
\end{aligned}
$$

∎

Now, recall that $x$ is the true parameter. For $a \neq a_x^*$, we have that:

$$
\begin{aligned}
T_n(a) &= \sum_{t=1}^{n} \mathbf{1}(I_t = a) = 1 + \sum_{t=K+1}^{n} \mathbf{1}(I_t = a) \\
&= 1 + \sum_{t=K+1}^{n} \mathbf{1}(I_t = a, C1(z) \text{ is satisfied at time } t \text{ for some } z \in \mathcal{X}) \\
&\quad + \sum_{t=K+1}^{n} \mathbf{1}(I_t = a, C2(z) \text{ is satisfied at time } t \text{ for some } z \in \mathcal{X}) \\
&\quad + \sum_{t=K+1}^{n} \mathbf{1}(I_t = a, C3 \text{ is satisfied at time } t) \\
&= 1 + \text{Term 1} + \text{Term 2} + \text{Term 3}.
\end{aligned}
\tag{2}
$$

Let us define $\mathcal{L}_a := \sup\{T_n(a) \geq 1 : \overline{\theta}_n(a) \notin \epsilon^*\text{-nbd}(\theta_x(a))\}$. Then by Lemma 1, $\mathbb{E}[\mathcal{L}_a] \leq \gamma(\epsilon^*)$. Moreover, we have that Term 1 $\leq \mathcal{L}_a$ (since $B(z)$ is empty in this case), and Term 3 $\leq \sum_{a \in \mathcal{A}} \mathcal{L}_a$. Therefore,

$$
\mathbb{E}[\text{Term 1}] \leq \gamma(\epsilon^*),
\tag{3}
$$

and

$$
\mathbb{E}[\text{Term 3}] \leq K\gamma(\epsilon^*).
\tag{4}
$$

Now, we note that Term 2 can be expanded as:

$$
\begin{aligned}
\text{Term 2} &= \sum_{t=K+1}^{n} \mathbf{1}(I_t = a, C2(z) \text{ is satisfied at time } t \text{ for some } z \neq x) \\
&\quad + \sum_{t=K+1}^{n} \mathbf{1}(I_t = a, C2(x) \text{ is satisfied at time } t) \\
&= \text{Term 2a} + \text{Term 2b}.
\end{aligned}
\tag{5}
$$

Let us consider the case when $B(x)$ is empty. Then we have that:

$$\mathbb{E}[\text{Term 2a}] = \sum_{t=K+1}^{n} \mathbb{E}[\mathbf{1}(I_t = a, C2(z) \text{ is satisfied at time } t \text{ for some } z)]$$

$$\leq \sum_{t=K+1}^{n} \mathbb{E}[\mathbf{1}(I_t = a_x^*, C2(z) \text{ is satisfied at time } t \text{ for some } z)]\mathbf{1}(a \in \mathcal{E})$$

$$\leq \gamma(\epsilon^*)\mathbf{1}(a \in \mathcal{E}) \qquad (\text{if } B(x) \text{ is empty}). \tag{6}$$

The first inequality of the above expression is due to the following fact: under $C2(z)$, the UCB algorithm is run over $\mathcal{E}$, which also includes $a_x^*$; and since $x$ is the true parameter, the expected time spent on $a_x^*$ is larger than the expected time spent on any other arm. The second inequality is due to the following fact: the condition $C2(z)$ is satisfied for some $z \neq x$, which means that $B(z)$ is non-empty; and since $B(x)$ is empty, it must be that $\theta_x(a_x^*) \neq \theta_z(a_x^*)$.

Now, if $B(x)$ is non-empty, then we have that:

$$\mathbb{E}[\text{Term 2a}] = \sum_{t=K+1}^{n} \mathbb{E}[\mathbf{1}(I_t = a, C2(z) \text{ is satisfied at time } t \text{ for some } z)]$$

$$\leq \left( \frac{8}{[\Delta_x(a)]^2} \ln n + \frac{\pi^2}{3} \right) \mathbf{1}(a \in \mathcal{E}) \qquad (\text{if } B(x) \text{ is non-empty}). \tag{7}$$

The above inequality is due to the UCB result in [2].

Finally, we have that:

$$\mathbb{E}[\text{Term 2b}] \leq \left( \frac{8}{[\Delta_x(a)]^2} \ln n + \frac{\pi^2}{3} \right) \mathbf{1}(a \in \mathcal{E}). \tag{8}$$

Combining (2)-(8) yields the result.

*C. The Unif - Clustering - UCB-ET($\delta$) algorithm*

---

**Algorithm 2** Unif - Clustering - UCB-ET($\delta$)

---

**Input:** $M_0$ (number of pilot users), $N$ (number of types), $K$ (number of arms), $\delta$ (parameter)

**Output:** $I_1, I_2, \ldots$ (indices of the arm to be pulled)

**Initialize:** $\bar{\bar{\theta}}_i \leftarrow 0$ for $i = 1, \ldots, N$ (estimated parameter vector for each type), $P \leftarrow \emptyset$ (pilot set)

**Initialize:** $\mu(u) \leftarrow 0$ for $u \in \mathcal{U}$ (empirical reward vector for each user)

$t = 1$ (time starts)

**while** $t \geq 1$ **do**

    Obtain the user's ID $u_t$

    **if** $u_t$ is a new user **then**

        **if** $|P| < M_0$ (number of pilot users is less than $M_0$) **then**

            $P \leftarrow P \cup \{u_t\}$ (add $u_t$ to the pilot set)

            $I_t \leftarrow$ a uniformly random chosen arm

        **else**

            $I_t \leftarrow$ UCB-ET($\delta$) with the estimated parameter vectors $\{\bar{\bar{\theta}}_i\}$

        **end if**

    **else**

        **if** $u_t \in P$ ($u_t$ is a pilot user) **then**

            $I_t \leftarrow$ a uniformly random chosen arm

        **else**

            $I_t \leftarrow$ UCB-ET($\delta$) with the estimated parameter vectors $\{\bar{\bar{\theta}}_i\}$

        **end if**

    **end if**

    Obtain the reward, update the empirical reward vector $\mu(u_t)$

    **if** $|P| \geq M$ and $u_t \in P$ **then**

        Do clustering on $\{\mu_u\}_{u \in P}$ to obtain the estimated parameter vectors $\{\bar{\bar{\theta}}_i\}_{i=1,\ldots,N}$

    **end if**

    $t \leftarrow t + 1$

**end while**

---

*D. The UCB - Clustering - UCB-ET($\delta$) algorithm*

---

**Algorithm 3** UCB - Clustering - UCB-ET($\delta$)

---

**Input:** $M_0$ (number of pilot users), $N$ (number of types), $K$ (number of arms), $\delta$ (parameter)

**Output:** $I_1, I_2, \ldots$ (indices of the arm to be pulled)

**Initialize:** $\bar{\bar{\theta}}_i \leftarrow 0$ for $i = 1, \ldots, N$ (estimated parameter vector for each type), $P \leftarrow \emptyset$ (pilot set)

**Initialize:** $\mu(u) \leftarrow 0$ for $u \in \mathcal{U}$ (empirical reward vector for each user)

$t = 1$ (time starts)

**while** $t \geq 1$ **do**

    Obtain the user's ID $u_t$

    **if** $u_t$ is a new user **then**

        **if** $|P| < M_0$ (number of pilot users is less than $M_0$) **then**

            $P \leftarrow P \cup \{u_t\}$ (add $u_t$ to the pilot set)

            $I_t \leftarrow$ UCB for user $u_t$ only

        **else**

            $I_t \leftarrow$ UCB-ET($\delta$) with the estimated parameter vectors $\{\bar{\theta}_i\}$

        **end if**

    **else**

        **if** $u_t \in P$ ($u_t$ is a pilot user) **then**

            $I_t \leftarrow$ UCB for user $u_t$ only

        **else**

            $I_t \leftarrow$ UCB-ET($\delta$) with the estimated parameter vectors $\{\bar{\theta}_i\}$

        **end if**

    **end if**

    Obtain the reward, update the empirical reward vector $\mu(u_t)$

    **if** $|P| \geq M$ and $u_t \in P$ **then**

        Do clustering on $\{\mu_u\}_{u \in P}$ to obtain the estimated parameter vectors $\{\bar{\theta}_i\}_{i=1,\ldots,N}$

    **end if**

    $t \leftarrow t + 1$

**end while**

---

*E. The Clustering & UCB Continuously algorithm*

---

**Algorithm 4** Clustering & UCB Continuously

---

**Input:** $N$ (number of types), $K$ (number of arms), $M_{th}$ (parameter)

**Output:** $I_1, I_2, \ldots$ (indices of the arm to be pulled)

**Initialize:** $\mu(u) \leftarrow 0$ for $u \in \mathcal{U}$ (empirical reward vector for each user)

**Initialize:** $U \leftarrow \emptyset$ (set of current users)

$t = 1$ (time starts)

**while** $t \geq 1$ **do**

    Obtain the user's ID $u_t$

    **if** $u_t$ is a new user **then**

        $U \leftarrow U \cup \{u_t\}$ (add new user)

        **if** $|U| < M_{th}$ (number of current users is less than $M_{th}$) **then**

            $I_t \leftarrow$ UCB for user $u_t$ only

        **else**

            Do clustering on $\{\mu_u\}_{u \in U}$ to obtain $N$ clusters

            $I_t \leftarrow$ UCB for the cluster that $u_t$ belonging to (using data from all users in that cluster)

        **end if**

    **else**

        **if** $|U| < M_{th}$ (number of current users is less than $M_{th}$) **then**

            $I_t \leftarrow$ UCB for user $u_t$ only

        **else**

            Do clustering on $\{\mu_u\}_{u \in U}$ to obtain $N$ clusters

            $I_t \leftarrow$ UCB for the cluster that $u_t$ belonging to (using data from all users in that cluster)

        **end if**

    **end if**

    $t \leftarrow t + 1$

**end while**

---